\documentclass{article}

\usepackage[preprint,nonatbib]{nips_2018}

\usepackage[utf8]{inputenc} 
\usepackage[T1]{fontenc}    
\usepackage{hyperref}       
\usepackage{url}            
\usepackage{booktabs}       
\usepackage{amsfonts,amsmath}       
\usepackage{nicefrac}       
\usepackage{microtype}      
\usepackage[algoruled,boxed,lined]{algorithm2e}
\usepackage{graphicx}
\usepackage{subfig}
\title{An Unsupervised Approach to Solving Inverse Problems using Generative Adversarial Networks}

\author{
  Rushil Anirudh \\
  Center for Applied Scientific Computing\\
  Lawrence Livermore National Laboratory\\
  \texttt{anirudh1@llnl.gov} \\
  \And
  Jayaraman J. Thiagarajan \\
  Center for Applied Scientific Computing \\
  Lawrence Livermore National Laboratory \\
  \texttt{jjayaram@llnl.gov} \\
  \AND
  Bhavya Kailkhura \\
  Center for Applied Scientific Computing \\
  Lawrence Livermore National Laboratory \\
  \texttt{kailkhura1@llnl.gov} \\
  \And
  Timo Bremer \\
  Center for Applied Scientific Computing \\
  Lawrence Livermore National Laboratory \\
  \texttt{bremer5@llnl.gov} \\
}

\begin{document}

\maketitle

\begin{abstract}
Solving inverse problems continues to be a challenge in a wide array of applications ranging from deblurring, image inpainting, source separation etc. Most existing techniques solve such inverse problems by either explicitly or implicitly finding the inverse of the model. The former class of techniques require explicit knowledge of the measurement process which can be unrealistic, and rely on strong analytical regularizers to constrain the solution space, which often do not generalize well. The latter approaches have had remarkable success in part due to deep learning, but require a large collection of source-observation pairs, which can be prohibitively expensive.  In this paper, we propose an unsupervised technique to solve inverse problems with generative adversarial networks (GANs). Using a pre-trained GAN in the space of source signals, we show that one can reliably recover solutions to under determined problems in a `blind' fashion, i.e., without knowledge of the measurement process. We solve this by making successive estimates on the model and the solution in an iterative fashion. We show promising results in three challenging applications -- blind source separation, image deblurring, and recovering an image from its edge map, and perform better than several baselines.


\end{abstract}

\section{Introduction}
A large class of machine learning techniques has been devoted to solving inverse problems that arise in different application domains. Formally, this refers to problems that take the form $Y = \mathcal{F}(X) + \mathbf{n}$, wherein the goal is to recover the true source $X$ from its noisy observation $Y$ \cite{ribes2008linear}. The mapping $\mathcal{F}$ includes a wide variety of corruption functions \cite{xie2012image,pathak2016context,yeh2017semantic}, measurement processes \cite{chang2017one,Reconnet,anirudh2017lose} and mixing models. In addition to its broad applicability, the highly under-determined nature of this formulation has made it an important problem among machine learning researchers for over two decades. In its most general form, this is posed as Maximum a Posteriori (MAP) estimation, and a common recurring idea that comes up in several existing solutions is to regularize the problem by restricting the space of solutions that $X$ can assume, through appropriate prior models. For example, \textit{sparsity} prior with respect to a latent basis has enabled effective recovery from compressed measurements of many natural signals \cite{thiagarajan2014image}. However, what differentiates between the variety of existing solutions is the degree to which they assume knowledge of $\mathcal{F}$ -- while most approaches assume access to exact parameterization of $\mathcal{F}$, others make more relaxed assumptions, e.g. distribution of $\mathcal{F}$ is known \cite{AsimDeblurGAN,bora2018ambientgan}. However, it is important to note that, even with a fully known $\mathcal{F}$, the adherence of observed data to the prior model can be insufficient, thus making the solutions highly non-robust.

In order to circumvent the challenge of choosing an appropriate prior for the data at hand, and dispense the need to know $\mathcal{F}$ in its analytical form, more recently, deep neural networks have been employed to directly build a surrogate for the inversion process $\hat{X} = \mathcal{T}(Y)$, where $\mathcal{T}$ is a neural network that implicitly approximates $\mathcal{F}^{-1}$; for example, some recent applications are in CS reconstruction \cite{chang2017one,Reconnet}, super-resolution \cite{ledig2016photo}, CT image reconstruction \cite{anirudh2017lose} etc. Such a supervisory approach relies on the ability to collect enough training data in the form of $(X, Y)$ pairs (works even for black-box $\mathcal{F}$). Despite the unprecedented success of this approach, in many scenarios it can be impractical to obtain such large training pairs. Another inherent limitation of this approach is that it does not provide an actual estimte of the mapping $\mathcal{F}$, which can be crucial in certain applications, e.g. blind source separation, or in imaging. Instead, we propose a general unsupervised solution for inverse problems that utilizes pre-trained generative models to form the prior and simple shallow networks to approximate the mapping $\mathcal{F}$, with the goal of effectively producing the observed data $Y \approx \hat{\mathcal{F}}(\hat{X})$.We show that, through an alternating optimization algorithm,  one can effectively recover both the unknown mapping as well as the true sources solely from a limited set of observations. This is conceptually similar to the problem of jointly learning sparse representations and an associated overcomplete dictionary from the data, where the two unknowns are alternatively inferred with a sparsity regularizer, e.g. Laplacian prior \cite{thiagarajan2011optimality}.

Deep generative models, such as variational autoencoders \cite{kingma2013auto} and Generative Adversarial Networks (GANs)\cite{GANGoodfellow}, have enabled non-parameteric inferencing of complex data distributions. Consequently, there has been a recent surge in utilizing data-driven generative models to sample from distributions in several traditionally difficult, unsupervised learning problems. In the context of inverse problems, a Generative Adversarial Network that models $P(X)$ can naturally act as a strong prior for $X$, thus eliminating the need to choose a prior that can regularize the problem while also being convenient for optimization. Furthermore, even with a shallow network, we are able to represent a large class of mappings $\mathcal{F}$ (both linear and simple non-linear mappings) thus allowing a robust approximation of the measurement process without the need for an analytical form or even a prior on $\mathcal{F}$. Consequently, the system remains reasonably agnostic to the current task being solved -- for example, the same system can be reused to solve a deblurring task and  reconstruct an image from its edgemap.

To the best of our knowledge, this is the first work to provide a single, unsupervised solution to different inverse problems. When applied to standard inverse imaging problems, we show that our unsupervised approach performs competitively against baselines with full knowledge of $\mathcal{F}$ and significantly outperforms other existing unsupervised approaches. Interestingly, the proposed approach can be applied to more challenging scenarios such as blind source separation, wherein there are multiple sources corresponding to each observation, and there can be multiple observations (each with a different mapping) corresponding to the same set of sources. Even in highly underdetermined scenarios where conventional approaches such as independent component analysis fail completely, i.e. number of observations are significantly lesser than number of sources, we observe that our algorithm recovers the underlying sources with high-fidelity.

\section{Proposed Approach}
In this section we describe the proposed unsupervised solution for inverse problems. As described in the previous section, it does not assume knowledge of $\mathcal{F}$ or analytical form of the regularizer. In the blind source separation case, we assume that the number of sources is known \textit{a priori}, though existing techniques can be used for its estimation \cite{thiagarajan2013mixing}.


\textbf{Background:} The goal of several ill-posed inverse problems, such as compressed recovery or source separation, is to estimate the true source signals from an underdetermined system of noisy measurements. In order to solve these ill-posed problems in a tractable fashion, one usually needs a prior knowledge on the structure of solution in the domain of signal $X$. In general, different prior assumptions lead to different forms of regularization,  using which several signal/image processing problems can be formulated as optimization problems and effectively solved using existing techniques.

Several successful approaches for solving inverse problems make two crucial assumptions: (a) parameters of the measurement process $\mathcal{F}$ are known; and (b) the prior on structure of the solution can be represented analytically (e.g., low rank or sparsity w.r.t. a known basis). Although widely used, these assumptions are very restrictive and hard to meet in several scenarios. While the first assumption simplifies the inversion process, it limits application to known corruption models. On the other hand, the latter assumption is targeted at simplifying the optimization process (e.g. convexity) such that existing techniques can be used effectively. For example, it has been widely known that natural images typically lie close to a collection of low dimensional sub-manifolds, however, a precise and analytical characterization for a given dataset is very challenging. Consequently, analytical approximations of the known low-dimensional structure (such as total variation regularization) is preferred, and convex optimization procedures are employed to solve this regularized alternative instead of the original recovery problem.

In order to overcome these challenges, we propose a unsupervised approach for solving inverse problems by (a) employing shallow neural networks to estimate the measurement process and, (b) utilizing deep generative models (GANs) as an effective, non-analytical regularizer. In the rest of this section, we formalize these ideas and describe the algorithm.

\noindent \textbf{Problem Formulation:}
Let us denote a set of $N$  observed measurements as $\mathbf{Y}^{obs}\in \mathbb{R}^{N \times K}$ where each column $Y_j^{obs} \in \mathbb{R}^{K}$ denotes an independent observation. For generality, we assume that each observed signal $Y_j^{obs}$ is composed using a set of $S$ source signals $\mathbf{X}_j \in \mathbb{R}^{S \times M}$, wherein we hypothesize that each of the source signals lie in a low-dimensional manifold $\mathcal{M}\subseteq \mathbb{R}^M$ and $K\leq M$. Note that, in the case of conventional inverse imaging problems such as de-blurring $S = 1$. Now, the sources and the observations are related through a unknown measurement process $\mathcal{F}$ by $Y_j^{obs} = \mathcal{F}\left(\mathbf{X}_j \right)+\mathbf{n}$.
Here $\mathbf{n}$ represents noise in the measurement process. The goal is to recover an estimate of the source signals $\hat{\mathbf{X}}_j$ from each of the $N$ observations. For the sake of simplicity, here we only consider a single observation for each set of sources, but it can be trivially generalized to multiple observations for scenarios such as blind source separation.


\noindent Several existing solutions for inverse problems assume $\mathcal{F}$ to be known and are formulated as:
\begin{equation}
\{\hat{\mathbf{X}}_j\}_{j = 1}^N = \underset{\{\mathbf{X}_j \in \mathbb{R}^{S \times M}\}_{j=1}^N}{\arg\min} \;
 \sum_{j = 1}^N \left\|Y_j^{obs}-\mathcal{F}(\mathbf{X}_j) \right\| + \lambda \mathcal{R}_\mathcal{M} (\mathbf{X}_j)
\end{equation}
where $\mathcal{R}_\mathcal{M}$ is an analytical regularizer (such as, $l_1$ norm, TV regularizer, etc.) and $\lambda$ is a regularization parameter. The hope of solving a regularized optimization problem is that regularization $\mathcal{R}_\mathcal{M}$ (if modeled precisely) will push the solution to lie on (or be near) the true signal manifold $\mathcal{M}$. As mentioned earlier, the above approach has serious limitations and motivates the proposed formulation as discussed next.

The proposed approach overcomes these modeling limitations by estimating both unknown mixing function and prior signal structure from the data itself. The unknown mixing function is parameterized by a neural network, denoted by $\hat{\mathcal{F}}$ acting as a surrogate, and prior signal structure is parameterized by using a GAN -- $\mathcal{G}: \mathbb{R}^{T} \mapsto \mathbb{R}^M$, where $\mathcal{G}$ denotes the generator in the GAN and $T$ is the number of latent dimensions (set to $100$ in our experiments). The implicit regularization with this ``GAN prior'' results in the following optimization problem:
\begin{equation}
\label{eq:obj}
\hat{\mathcal{F}}^*, \{\mathbf{z}^*_j\}_{j = 1}^N= \underset{\hat{\mathcal{F}},\{\mathbf{z}_j\in \mathbb{R}^{S \times T}\}_{j=1}^N}{\arg\min} \;
\sum_{j = 1}^N \left\|Y_j^{obs}-\hat{\mathcal{F}}(\mathcal{G} (\mathbf{z}_j))\right\|,
\end{equation}where we have $\hat{\mathbf{X}}_j = \mathcal{G}(\mathbf{z}_j)$. Since we have two unknowns in \eqref{eq:obj}, we employ an alternating optimization which allows us to find the optimal $\hat{\mathcal{F}}^*$ and $\{\mathbf{z}^*_j\}_{j=1}^N$. In particular, since $\mathcal{G}$ and $\hat{\mathcal{F}}$
are differentiable, we can evaluate the gradients of the objective in \eqref{eq:obj}, using backpropagation and use existing gradient based optimizers. In addition to computing gradients with respect to $\mathbf{z}_j$, we also perform clipping in order to restrict it within the desired range (like $[-1,1]$) resulting in a projected gradient descent optimization.
With sufficiently large number of observations $N$, the measurement process $\mathcal{F}$ can be estimated with high fidelity (as demonstrated in our experiments). As the deep generative model is obtained via unsupervised training on the data directly, it can very precisely model complex data distributions. Consequently, this enables us to utilize the low-dimensional structural information from data manifolds which cannot be otherwise modeled analytically with conventional regularizers. Furthermore, by providing a surrogate for the measurement process, it dispenses the limitation of supervised approaches that map directly from observations to the source signals.

\noindent Other applications like blurring and edge maps can be considered special cases of the problem formulation in \eqref{eq:obj}, with $S=1$, and $\mathcal{F}$ acts as a linear operator on $\mathbf{X}$.

\noindent\textbf{Algorithm}
The algorithm to perform the alternating optimization is shown in Algorithm \ref{mainAlg}. We run the inner loops for updating the surrogate and the latent parameters of $\mathcal{G}$ for $T_1$ and $T_2$ iterations respectively. The projection operation denoted by $ \mathcal{P} $ is the clipping operation, where we restrict the update on $\mathbf{z}_j$ to lie within the desired range. In addition to the reconstruction error, we also incorporate a perceptual loss, that penalizes unrealistic images. For a given discriminator model $\mathcal{D}$ and a generator $\mathcal{G} $, this loss is given by: $\mathcal{L}_{per} = \sum_{j=1}^N\sum_{i=1}^S \log(1-\mathcal{D}(\mathcal{G}(z_j^i)))$, where $z_j^i$ is the $i^{\text{th}}$ column of the latent matrix $\mathbf{z}_j$. Note, this is the same as the generator loss of the pre-trained GAN.

\noindent An advantage of the algorithm described here is that it only depends on $\mathbf{Y}^{obs}$ to perform the update. As a result, not only does it not require any paired training data to be collected, but the procedure also lends itself to a task-agnostic inference wherein the user does not need to specify  \emph{a priori} what the current task being solved is. This is in contrast to most existing deep learning based solutions today, which are optimized to solve a specific task using training data collected in advance.

\begin{algorithm}[t]
\SetKwFunction{GAN}{GAN}
\SetKwFunction{random}{random}
\SetKwInOut{Input}{input}
\SetKwInOut{Output}{output}
\Input{Number of sources $ S$, Observations $\mathbf{Y}^{obs} \in \mathbb{R}^{N\times K}$, Pre-trained GAN $\mathcal{G}$.}
\Output{Estimated sources $\{\hat{\mathbf{X}}_j \in \mathbb{R}^{S \times M}\}_{j=1}^N$, Surrogate model $\hat{\mathcal{F}}$ }
\BlankLine
For $j = 1, \cdots, N$, initialize $\mathbf{z}_j^{(0)} \in \mathbb{R}^{S \times T}$ randomly using an uniform distribution $\mathcal{U}(-1,1)$ .
\BlankLine
$\mathbf{z}_j^* = \mathbf{z}_j^{(0)} ~ \forall j$; //initial best guess is random

Initialize $\hat{\mathcal{F}}$ as a shallow neural network with random weights $\mathbf{\Theta}^{(0)}$.
\BlankLine

\For{$t\leftarrow 0$ \KwTo $T$}{
    \BlankLine

  \For{$t_1\leftarrow 0$ \KwTo $T_1$}{
    $Y_j^{est} \leftarrow \hat{\mathcal{F}}\left(\mathcal{G}(\mathbf{z}_j^*); \mathbf{\Theta}^{(t_1)}\right),~\forall j$; // present best guess of sources

    $\mathcal{L} = \sum_{j = 1}^N \|Y_j^{est} - Y_j^{obs}\| + \alpha\mathcal{L}_{per}$;

    $\mathbf{\Theta}^{(t_1+1)} \leftarrow \mathbf{\Theta}^{(t_1)} - \lambda_1~\nabla_\mathbf{\Theta}(\mathcal{L}) $; // gradient descent
    }
    $\mathbf{\Theta}^* = \mathbf{\Theta}^{(T_1)}$

    \For{$t_2\leftarrow 0$ \KwTo $T_2$}{
      $Y_j^{est} \leftarrow \hat{\mathcal{F}}\left(\mathcal{G}(\mathbf{z}_j^{(t_2)});\mathbf{\Theta}^*\right),~ \forall j$; // present best guess of measurement process

      $\mathcal{L} = \sum_{j = 1}^N \|Y_j^{est} - Y_j^{obs}\| + \alpha\mathcal{L}_{per}$;

      $\tilde{\mathbf{z}}_j^{(t_2+1)} \leftarrow \mathbf{z}_j^{(t_2)} - \lambda_2~\nabla_\mathbf{z}(\mathcal{L}),~ \forall j$; // gradient descent

      $\mathbf{z}_j^{(t_2+1)} \leftarrow \mathcal{P}\left(\tilde{\mathbf{z}}_j^{(t_2+1)}\right)~ \forall j$; //projection operation
      }
      \BlankLine
      $\mathbf{z}_j^* = \mathbf{z}_j^{(T_2)}, ~\forall j$

  }
  return $\hat{\mathcal{F}}^*$, $\hat{\mathbf{X}}_j = \mathcal{G}(\mathbf{z}_j^*),~\forall j$.
\caption{Proposed Algorithm}\label{mainAlg}
\end{algorithm}

\section{Experiments}
In this section, we demonstrate the effectiveness of our proposed solution using three scenarios that rely on solving highly ill-posed inverse problems -- (a) image deblurring, (b) recovering an image from its edge map, and (c) blind source separation with unknown mixing configuration. In all three scenarios, we assume no knowledge of the measurement process, and attempt to estimate the mapping $\mathcal{F}$ and true source signals jointly. Consequently, with sufficient number of observations and under suitable assumptions on the capacity of the shallow network used  to approximate $\mathcal{F}$, our inversion system can be directly reused across different inverse problems.

\subsection{Inverting convolutional operators -- blur and edge transforms}
In this experiment, we attempt to recover images after they have been filtered with a convolutional kernel such as a blur or an edge kernel. These belong to a large class of inverse problems commonly characterized by convolutional filtering operations. The convolution operation can also be interpreted as matrix multiplication when the convolution kernel is represented in its Toeplitz matrix form, with no requirement to be full rank.

 \noindent For the deblurring experiment, we used a standard $20 \times 20$ Gaussian blur kernel on face images, with a scale parameter of $5$, which is severe enough to ensure that no facial features are discernable from the transformed images. For the edge map experiment, we used a $3\times 3$ edge kernel given by$\left[ {\begin{array}{ccc}
   1 & 0 & -1\\
   2 & 0 & -2\\
   1 & 0 & -1\\
  \end{array} } \right]$.

\paragraph{Experimental Setup:}
First, we build the prior model by training a DCGAN \cite{radford2015unsupervised} on the CelebA dataset \cite{liu2015faceattributes},  which consists 202,599 images of which we use $90\%$ for training, and used the remaining to perform our experiments. To train the DCGAN, we used the hyper-parameters suggested in \cite{radford2015unsupervised}. As described in Section 2, we model the surrogate $\hat{\mathcal{F}}$ using a 2-layer convolutional network without any non-linearities, since the transformations considered are linear. In order to speed up convergence and avoid overfitting, we used a small set of filters -- $16$ filters of size $5\times 5\times 3$ in the first layer, followed by $3$ filters of size $5 \times 5\times 16$. We also experimented with non-linearities such as ReLU, but found convergence to be much faster without them for the linear inverse problems considered in this experiment.

\noindent In general, we observe that a complex surrogate with larger number of parameters in $\hat{\mathcal{F}}$, requires a larger set of observations in order to avoid over-fitting. In this experiment, we found that even just $25-50$ observations, i.e., images transformed using the same linear model, are sufficient to train an effective surrogate. Even though this is an extremely small batch size, we believe the network does not overfit easily because the images used to train the surrogate are updated constantly with the GAN as outlined in Algorithm \ref{mainAlg}. 

In the alternating optimization, we first update $\hat{\mathcal{F}}$ for $T_1 = 50$ iterations, followed by updating the sources $\{\hat{\mathbf{X}}_j\}$ by optimizing over the latent variables of the GAN model for $T_2=50$ iterations. The optimization procedure typically achieves convergence in about $T = 100$ epochs. We used the Adam Optimizer \cite{kingma2014adam} for both the updates, with a learning rate of $4\mathrm{e}{-3}$ to update the surrogate and a slightly lower rate of $3\mathrm{e}{-4}$ for updating the latent parameters of the generative model. In addition, we clip the updates on $\{\mathbf{z}_j\}$ (see \ref{mainAlg}) at each step.

\begin{figure*}[!t]
	\centering
	\subfloat[Inverting the blur operation]{
		\includegraphics[width=.85\linewidth]{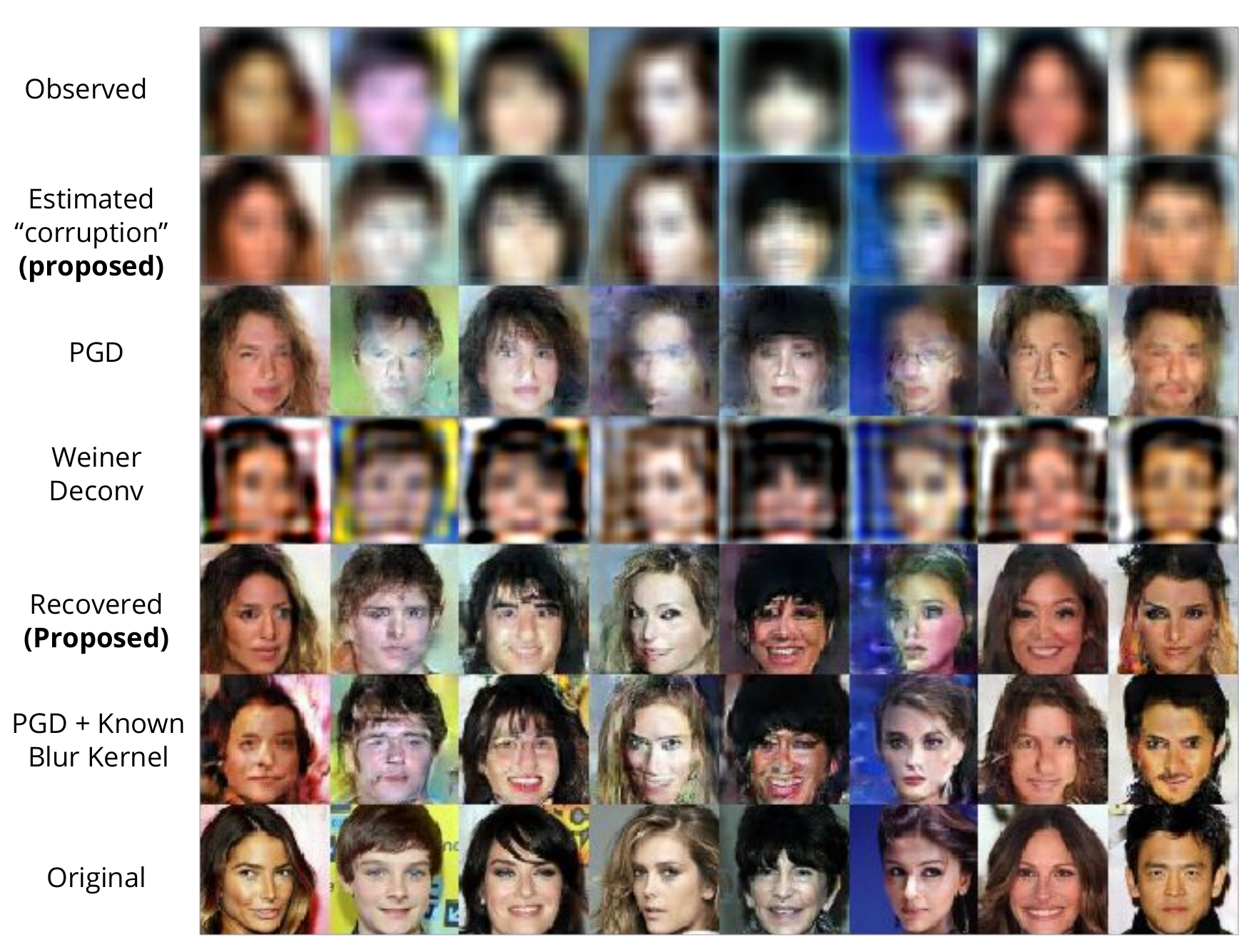}
		\label{fig:deblur}}

	\subfloat[Inverting the edgemap operation]{
		\includegraphics[width=.9\linewidth]{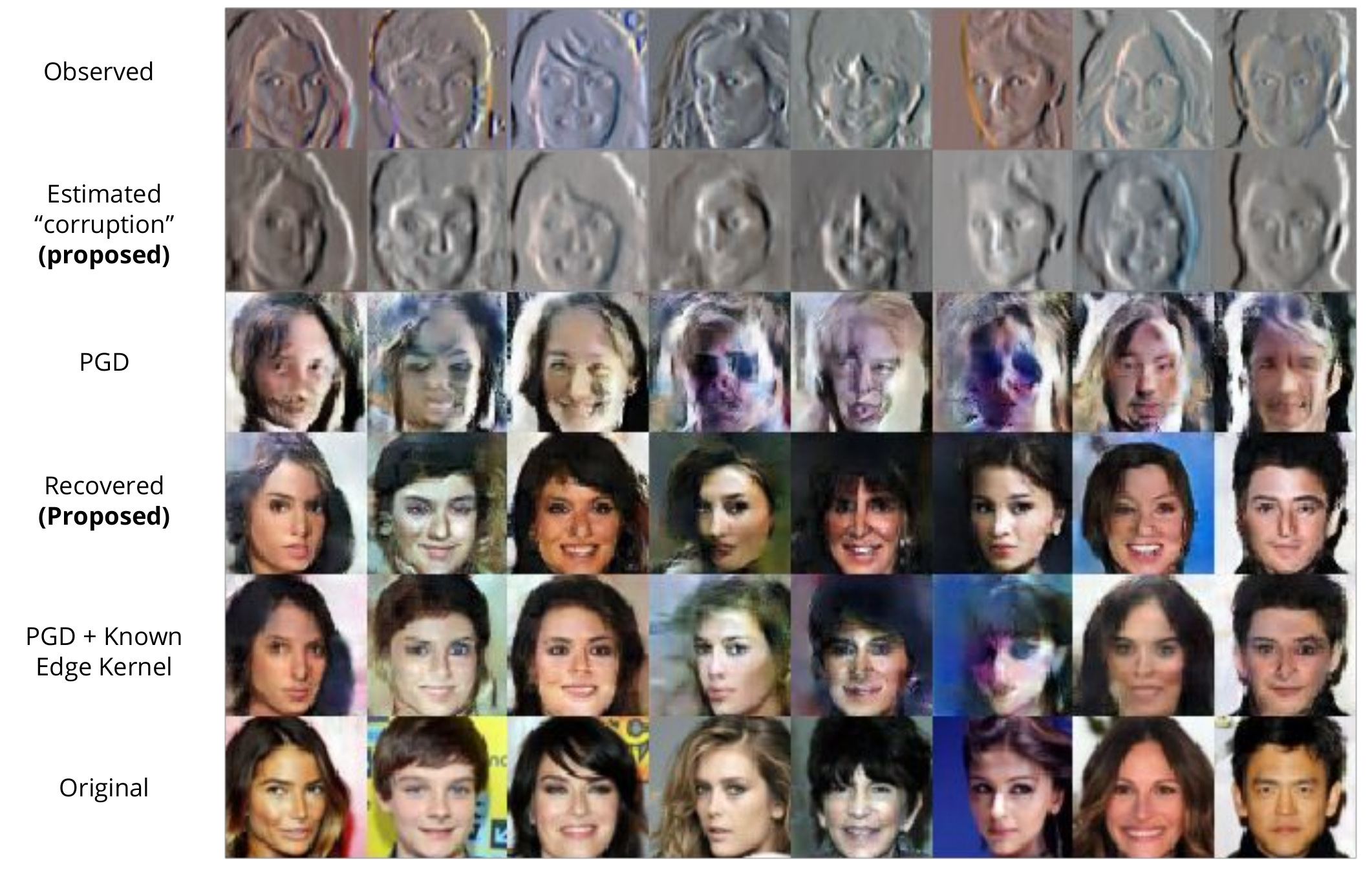}
		\label{fig:deedge}}
	\caption{\small{Blind inversion of the blur filter and the edge filter using the proposed approach. It should be noted that in both the cases the system remains unchanged, i.e., the algorithm does not need to know if its being provided with an edge map or a blured image. It is observed in figure \ref{fig:deblur}, that most facial attributes are removed, yet we estimate the most likely original image accurately.}}
	\label{fig:conv}
\end{figure*}

\textbf{Losses:} As noted in several existing image recovery efforts, such as \cite{yeh2017semantic}, we also observed that an $\ell_1$ loss worked better in image recovery problems than the $\ell_2$ loss. In addition, following \cite{yeh2017semantic}, we also penalized images that are not ``perceptually'' meaningful, i.e., encourage images that reduce the generator loss from the pre-trained GAN. We found the balance between these two losses is highly sensitive and governed by the weight $\alpha$, and observed $\alpha = 1\mathrm{e}{-4}$  to be a suitable choice.

\textbf{Results:}
We compare the proposed approach with the following baselines: (a) \emph{PGD}: Use projected gradient descent directly with respect to the loss $\mathcal{L} = \sum_j \|\mathcal{G}(\mathbf{z}_j) - \mathbf{Y}_j^{obs}\|+ \mathcal{L}_p$ , while dispensing entirely, the use of a measurement process. This simple baseline works reasonably well for simple transormations, e.g. blur; (b) \emph{Weiner Deconv}: For the blur operation, we use traditional signal processing restoration based on Weiner deconvolution; (c) \emph{PGD + Known $\mathcal{F}$}: This is representative of most existing GAN-based recovery approaches such as  \cite{yeh2017semantic, bora2017compressed,shah2018solving} etc. where the exact parameterization of $\mathcal{F}$ is known.

We observe that the proposed approach significantly outperforms the simpler baselines that do not have access to $\mathcal{F}$, while performing competitively to the cases where it is known. Results for test cases on the CelebA dataset are shown for deblurring in Figure \ref{fig:deblur}, and edgemap to image  recovery in Figure \ref{fig:deedge}. Furthermore, in both cases, we observe that the proposed approach produces an effective surrogate to produce the actual observations.

\subsection{Blind source separation}
Next, we study the application of blind source separation (BSS), commonly encountered in signal and image processing, where the goal is the following: recovering individual sources when observing only a linear or non-linear mixture of these sources. We fix $S=3$ in our experiments. This problem is severly under-determined when the number of observations is smaller than the number of sources, and it can be over-determined when number of observations is higher. Furthermore, most traditional techniques assume $\mathcal{F}$ to be linear in order to solve BSS, whereas our method is easily generalizes to even non-linear mixtures. In addition to the mixture observations, the proposed technique requires $S$, the number of sources as input. We model the mixing process using a fully connected 2-layered neural network with 16 units in the first layer, and ReLu activation after the first layer, followed by $N$ (number of observations) units in the final layer. Our model estimates the sources \emph{without training}, i.e., it behaves as an iterative algorithm at test time and can automatically work with any kind of mixing process.

\begin{figure*}[!htb]
\centering
\subfloat[Single observation, three sources. WHile there are no unique solutions to this problem, our approch finds highly likely solutions. The images are intensity normalized.]{
\includegraphics[width=.85\linewidth]{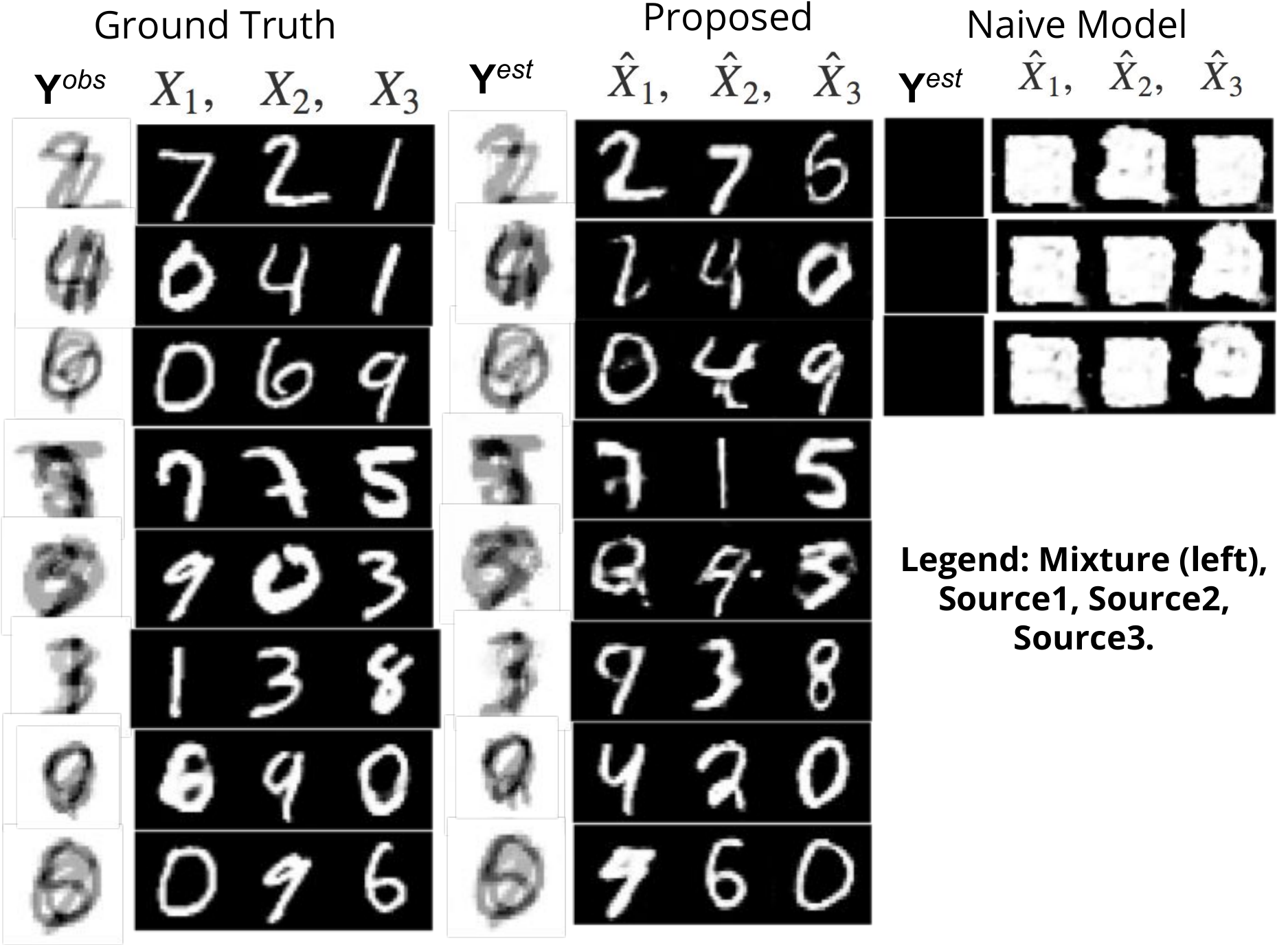}
\label{fig:bss1}}

\subfloat[Four observations, three sources.]{
\includegraphics[width=.85\linewidth]{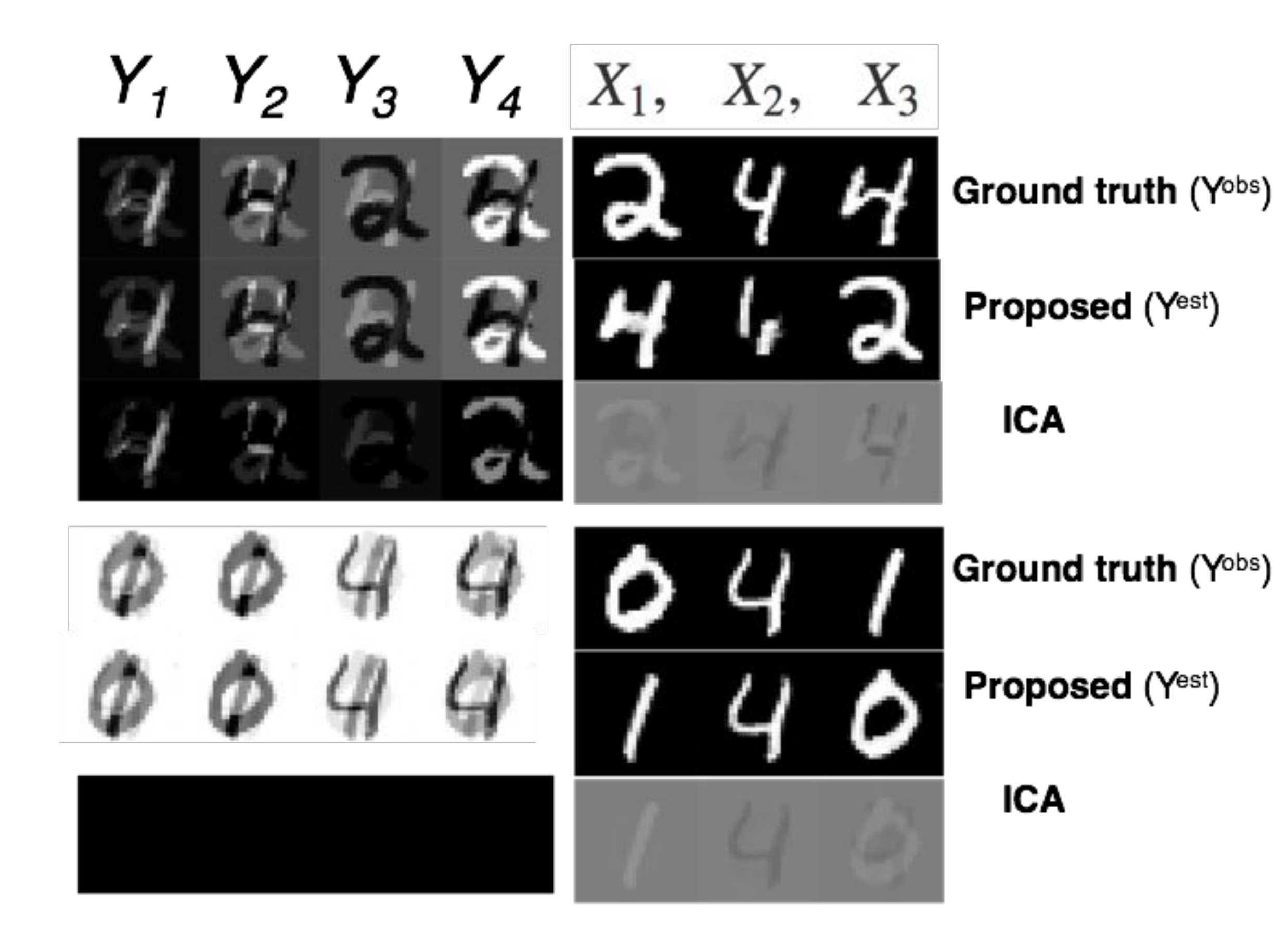}
\label{fig:bss2}}
\caption{\small{Blind Source Separation with single and multiple observations on the MNIST dataset. Images are intensity normalized.}}
\label{fig:bss}
\end{figure*}

\paragraph{Experimental Setup:} We pre-train a GAN with CNNs on the MNIST dataset \cite{lecun1998mnist} with 6 layers in the generator and discriminator; we also rescale the images to lie in the range of $[-1,1]$. Next, we simulate the mixing process using weights drawn from a random normal distribution, $\mathbf{M} \sim \mathcal{N}(-0.5,0.5)$, allowing for negative weights, where the final mixed observation is given by $Y_j^{obs} = |\mathbf{M}^T \mathbf{X}_j|$. We train each block of algorithm \ref{mainAlg} for $50$ iterations each, with $T = 100$. Our formulation can naturally handle scenarios with multiple observations as well, which is done by generalizing $\mathbf{M}$ to have multiple rows. We show results for blind source separation in Figure \ref{fig:bss}, for two distinct cases -- (a) \textit{under determined}: single non-linear mixture from three sources, and (b) \textit{over determined}: $4$ distinct non-linear mixtures from $3$ sources.

\paragraph{Results:}
We compare the performance of our approach with the following baselines: (a) \emph{Na\"ive Additive Model:} We assume a simple non-weighted additive model, such that the loss is given by $\mathcal{L} = \sum_j \|\Sigma_i^S \mathcal{G}(z_j^i) - Y_j^{obs}\| + \alpha\mathcal{L}_p$. We find this baseline to be surprisingly strong even for weighted additive mixtures, however it completely fails with non-linearities as shown in figure \ref{fig:bss1}; (b) \emph{Independent Component Analysis (ICA)}: Though ICA fails completely (not shown) for the under determined case, it fairs better when the number of observations is increased. However, the as shown in figure \ref{fig:bss2} it is highly inferior to the proposed approach. We see in figure \ref{fig:bss1}, that our approach is particularly effective in under determined scenarios, where we have a single observation composed from multiple sources. We clearly observe that sampling from a GAN is an effective way to address the blind source separation problem.

\section{Related Work}
With their ability to effectively model high-dimensional distributions, GANs \cite{GANGoodfellow} have enabled a large number of unsupervised methods for image-related inverse problems. Yeh et al. \cite{yeh2017semantic} showed that projected gradient descent (PGD) can be used to fill up arbitrary holes in an image in a semantically meaningful manner. More recently, several efforts have pursued solving inverse problems using PGD, for example in compressive sensing \cite{bora2017compressed, shah2018solving}, and deblurring \cite{AsimDeblurGAN}. Asim et. al \cite{AsimDeblurGAN} proposed an unsupervised technique for deblurring that is closely related to our work, where they use two separate GANs -- one for the blur kernel, and another for the images, to solve for the inverse. Ours differs from this technique in that we make fewer assumptions on the functions that can be recovered, and only require a single GAN on the images. In addition, it is not evident how their method generalizes to other challening inverse problems like blind source separation. AmbientGAN \cite{bora2018ambientgan}, is another related technique that allows one to obtain a GAN in the original space given its lossy measurements, which helps make our case stronger as it provides a way to train a GAN even without the original images. However, to solve the inverse problem, they still assume the \emph{form} of corruption or mesurement (for example as a random binary mask or a measurement matrix etc.), whereas we parameterize it as a neural network that needs to be trained. Finally, our work leverages the notion of a ``GAN prior" \cite{yeh2017semantic,shah2018solving, AsimDeblurGAN, bora2017compressed} -- the idea that optimizing in the latent space of a pre-trained GAN provides a powerful prior to solve several traditionally hard problems. As GANs become better, we expect new capabilities in solving challenging inverse problems to emerge.

\section{Discussion}
In this paper, we presented a proof of concept for unsupervised techniques to solve commonly encountered ill-conditioned inverse problems. By leveraging GANs as priors, we are able to recover solutions from blurred images, edge maps, and separate sources from underdetermined non-linear mixtures. A crucial observation is that this approach does not require knowledge of the task that is being solved. Assuming  that the true source was realized from a known distribution (approximated by the GAN), our method is able to identify what corruption/transformation the source signal has gone through, while also identifying the signal itself.

The proposed technique opens up several new possibilities for inverse problem recovery, before which some key aspects need generalization. First, a batch version of the current technique can enable training more complex functions, that require many more observations than those considered here. Next, the choice of the surrogate model $\hat{\mathcal{F}}$ dictates the type of functions that can be recovered. As with supervised training, more complex functions require more observations. Further, recent work \cite{ulyanov2017deep} has shown that the choice of architecture places an implicit prior on the family of functions that can be modeled, so for example using a convolutional neural net may limit learning certain kinds of functions like inpainting-masks.

\small{
\bibliographystyle{ieee}
\bibliography{refs}

\subsubsection*{Acknowledgement}
This work was performed under the auspices of the U.S. Department of Energy by Lawrence Livermore National Laboratory under Contract DE-AC52-07NA27344.

\subsubsection*{Disclaimer}
This document was prepared as an account of work sponsored by an agency of the United States government. Neither the United States government nor Lawrence Livermore National Security, LLC, nor any of their employees makes any warranty, expressed or implied, or assumes any legal liability or responsibility for the accuracy, completeness, or usefulness of any information, apparatus, product, or process disclosed, or represents that its use would not infringe privately owned rights. Reference herein to any specific commercial product, process, or service by trade name, trademark, manufacturer, or otherwise does not necessarily constitute or imply its endorsement, recommendation, or favoring by the United States government or Lawrence Livermore National Security, LLC. The views and opinions of authors expressed herein do not necessarily state or reflect those of the United States government or Lawrence Livermore National Security, LLC, and shall not be used for advertising or product endorsement purposes.
}

\end{document}